\documentclass[letterpaper, 10 pt, conference]{ieeeconf}  

\IEEEoverridecommandlockouts                              

\usepackage{graphicx}
\usepackage{amsmath}
\usepackage{amssymb}
\usepackage{wrapfig}
\usepackage{placeins}
\usepackage{hyperref}

\usepackage{tabu}
\usepackage{subfig}
\usepackage{booktabs}

\usepackage{array}
\newcolumntype{P}[1]{>{\centering\arraybackslash}p{#1}}

\title{\LARGE \bf
S-BEV: Semantic Birds-Eye View Representation for Weather and Lighting Invariant 3-DoF Localization
}

\author{Mokshith Voodarla$^{1}$, Shubham Shrivastava$^{2}$, Sagar Manglani$^{2}$, Ankit Vora$^{3}$, \\ Siddharth Agarwal$^{3}$, Punarjay Chakravarty$^{2}$%

\thanks{$^{1}$ Work done as an intern for Ford Greenfield Labs, affiliated with University of California, Berkeley.
        {\tt\small mvoodarla@berkeley.edu}}%
\thanks{$^{2}$ The authors are with Ford Greenfield Labs, Palo Alto, CA, USA.
        {\tt\small sshriva5@ford.com}, {\tt\small smanglan@ford.com}, {\tt\small pchakra5@ford.com}}%
\thanks{$^{3}$ The authors are with Ford AV LLC, Dearborn, MI, USA.
        {\tt\small avora3@ford.com}, {\tt\small sagarw20@ford.com}}%
}

\begin{document}

\maketitle
\thispagestyle{empty}
\pagestyle{empty}

\begin{abstract}
We describe a light-weight, weather and lighting invariant, Semantic Bird's Eye View (S-BEV) signature for vision-based vehicle re-localization. A topological map of S-BEV signatures is created during the first traversal of the route, which are used for coarse localization in subsequent route traversal. A fine-grained localizer is then trained to output the global 3-DoF pose of the vehicle using its S-BEV and its coarse localization. We conduct experiments on vKITTI2 virtual dataset and show the potential of the S-BEV to be robust to weather and lighting. We also demonstrate results with 2 vehicles on a 22 km long highway route in the Ford AV dataset.


\end{abstract}

\section{INTRODUCTION}
Today’s autonomous vehicles have the ability to safely drive in most urban environments during regular day time conditions. However, most vehicles encounter the exact same problem in which the sensor data received is significantly affected by weather, lighting, or other natural effects of the scene. Specifically, this affects the vehicle’s ability to accurately localize to a 3-DoF pose in the environment which hinders its ability to safely navigate to a given destination.

LiDAR tends to be expensive (cost and compute-wise) and still experiences issues of fewer detected points and greater noise in heavy weather conditions such as rain, snow, etc. On the other hand, cameras are cheap (cost and compute-wise) and collect data which can be more easily leveraged to be weather-invariant. Being less compute intensive also allows for implementation on smaller platforms such as delivery robots.

\begin{figure}
    \centering
    \includegraphics[scale=0.2]{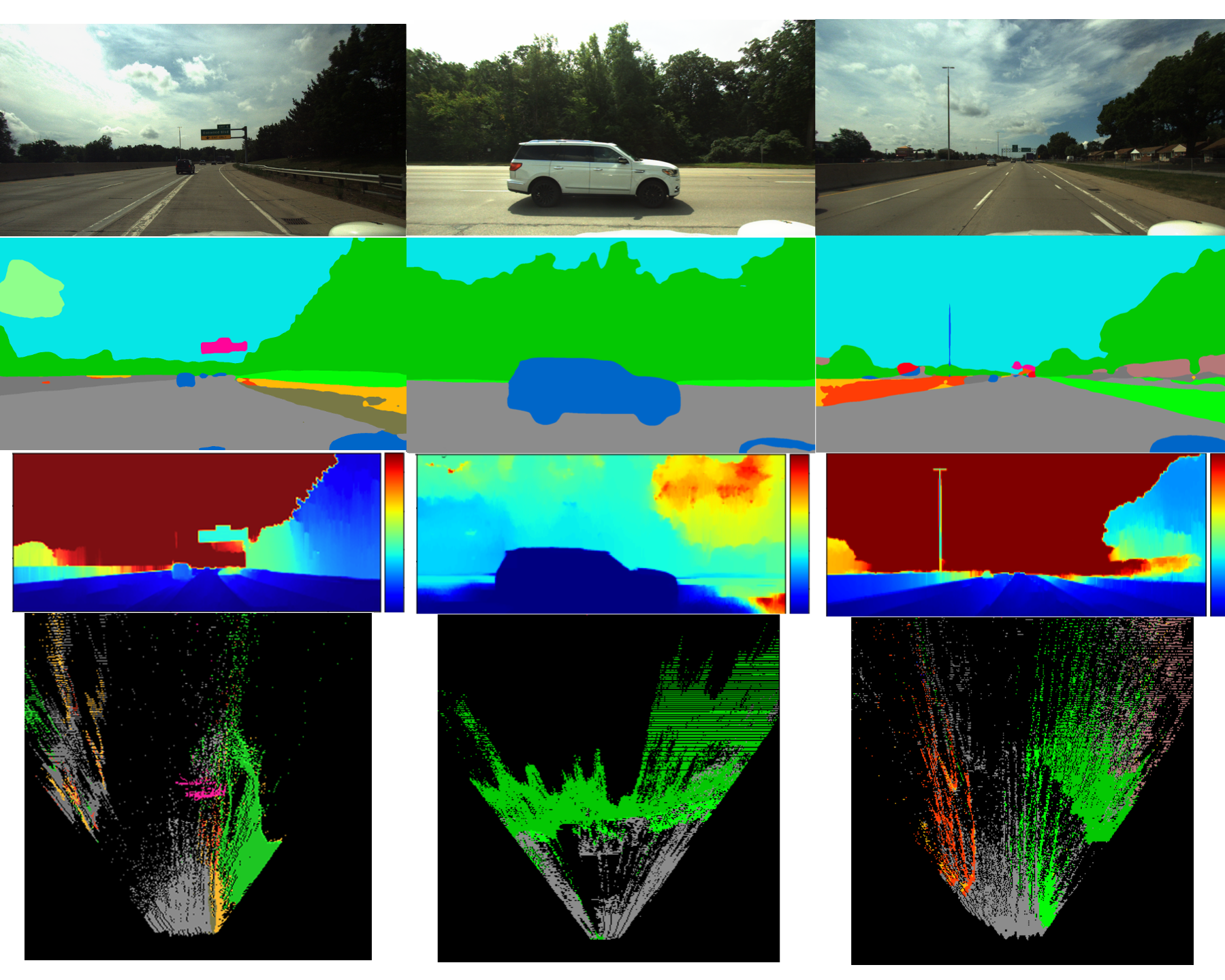}
    \caption{RGB images, segmentation images, depth maps and S-BEV representations from 3 topo-nodes in the Ford AV dataset. These S-BEVs are used for 3-DoF localization of the vehicle to a map.}
    \vspace*{-0.75cm} 
    \label{fig:overview}
\end{figure}

\begin{figure}[t]
\begin{center}
    \includegraphics[width=0.98\linewidth]{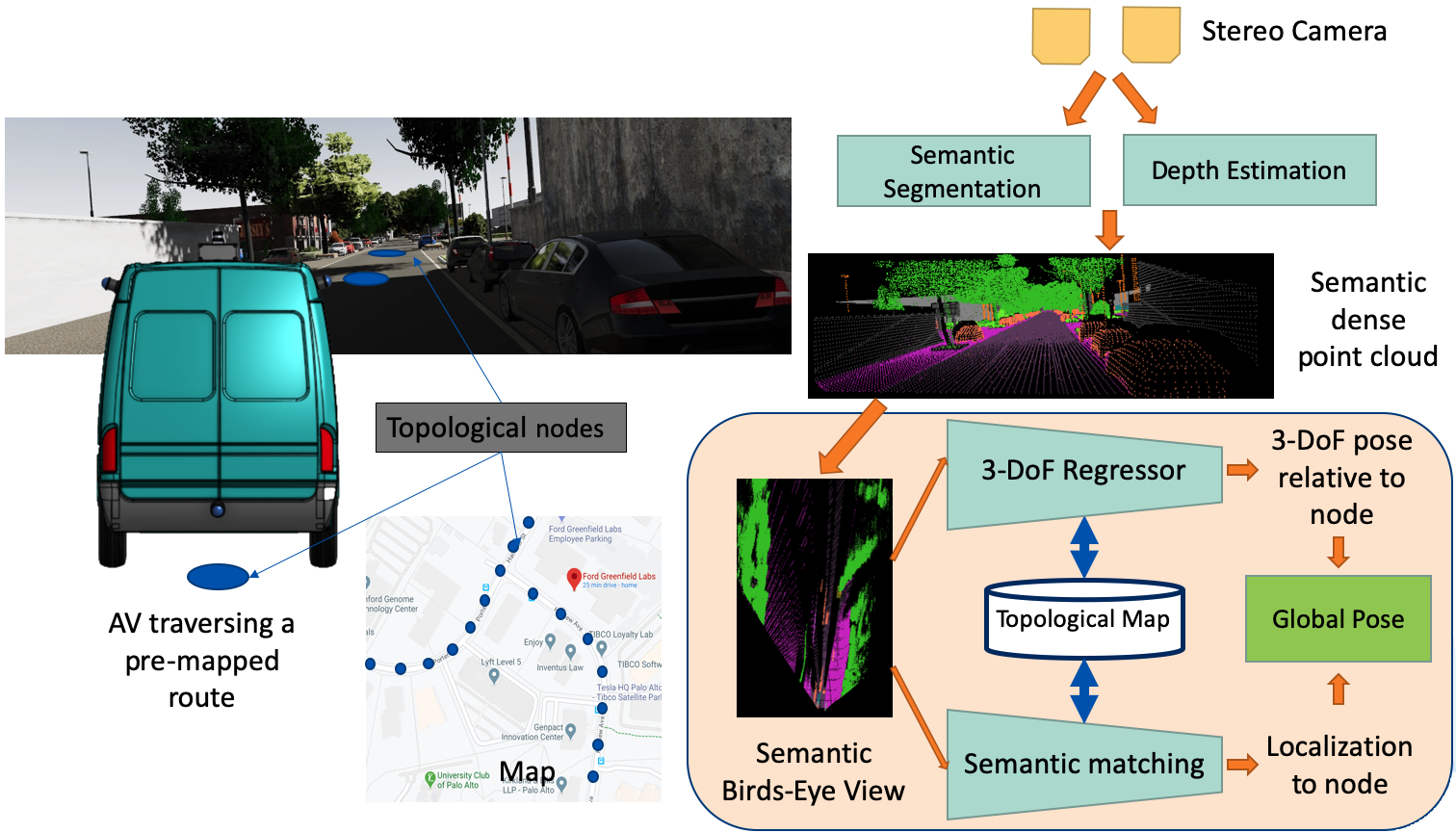}
\end{center}
\vspace*{-0.35cm} 
   \caption{S-BEV Localization System}
\label{fig:highLevelArch0}
\vspace*{-0.45cm} 
\end{figure}

\begin{figure}[t]
\begin{center}
    \includegraphics[width=0.98\linewidth]{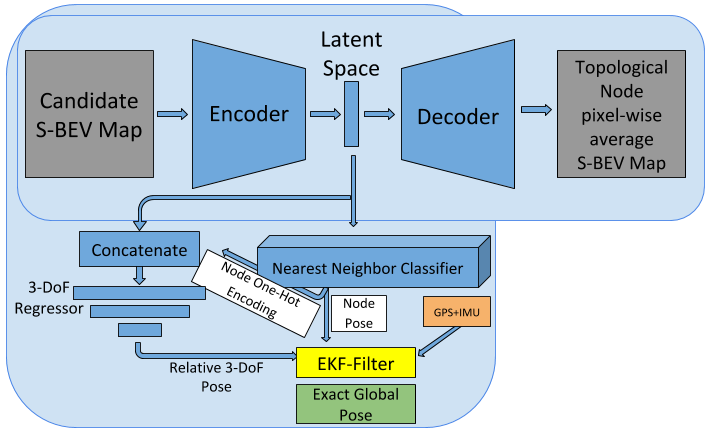}
\end{center}
\vspace*{-0.35cm} 
   \caption{S-BEV Localization Architecture}
\label{fig:highLevelArch}
\vspace*{-0.45cm} 
\end{figure}

We describe a lightweight method (see Figure \ref{fig:highLevelArch0} and \ref{fig:highLevelArch}) of leveraging stereo pair camera data to localize to an accurate 3-DoF position in any environment in varying weather conditions, light changes, and other natural effects such as snow and boulders. Instead of regular RGB input to our deep learning models, we construct a semantic birds-eye view (S-BEV) map as seen in Figure \ref{fig:overview}.


We are motivated by two properties of our semantic birds-eye view map. First, the semantic nature of our birds-eye view map is theoretically unaffected in lighting and weather conditions since semantic classes such as car, road, board, etc stay the same across lighting and seasons. Of course, this depends on the quality of the segmentation model but we show that most off-the-shelf segmentation models suffice as the segmenter for this pipeline. Second, a birds-eye view map characteristically doesn’t change much in shape or characteristics even when parts of a scene change such as mounds of snow or puddles of rain. While this may affect the RGB image, viewing the scene top-down reflects little to no change when situations like these are faced.

Our localizer begins with an autoencoder we use to encode information about a BEV map and its relative position to a topological node. During test time, we detach the decoder part of this autoencoder and perform a nearest neighbor classification of the computed latent embeddings between the candidate S-BEV and the set of previously mapped and collected S-BEVs. We then concatenate this classification with the latent embedding of the candidate image and feed it into a 3-DoF regressor which is built on fully connected layers and outputs a 3-dimensional vector representing the lateral translation, the longitudinal translation, and the angular difference between the closest topological node and the candidate BEV map. We then perform pose concatenation between the relative pose and the global pose of the closest node to find the global pose of the candidate image in the scene. The system architecture is shown in Figure \ref{fig:aeArch}.

Our system is trained with an equal amount of S-BEV samples per topological node. In most cases, scenes are unbalanced in terms of the number of samples which exist within nodes due to various factors such as travelling speed, traffic, etc. We undersample to have an equal amount of samples per node.

We suggest that this method of localization is a useful solution for robots and cars which operate in pre-mapped environments which could drastically change in lighting and weather in various times of day or various seasons in a year. Using an S-BEV map allows for the sole use of 2D Convolutional networks rather than 3D which contributes to increase in speed. Localizing to a topo-node and training the neural net to predict a relative pose also constrains the maximum error to within a topo-node. Our solution aims to have the right balance of robustness, accuracy and speed for deployment on relatively small and light-weight embedded systems. 

\section{RELATED WORK}
\subsection{Traditional Feature-based Localization}
Many existing large-scale visual localization approaches use local features such as SIFT \cite{SIFT} to create 2D-3D matches from 2D points in the image and 3D points in the 3D SfM model \cite{irschara2009pcls, toft2018semantic, svarm2017city}. These feature correspondences are then used to retrieve camera pose. While these tend to accurate attain accurate poses, they don't seem to generalize in terms of long-term season-by-season localization. They are also compute intensive due to their exhaustive matching and as the model grows in size, matching becomes less accurate due to perceptual aliasing where multiple places look the same. Situations like day and night are also a specific point struggle for this type of model because SIFT simply doesn't consistently match features in images taken at the same location in different weather or lighting conditions. While some approaches try to mitigate perceptual aliasing through a separate step for image retrieval \cite{irschara2009pcls} or through geometric outlier filtering \cite{zeisl2015voting, sattler2017outliers, svarm2017city}, they fail due to not enough visual or structural overlap existing between varying viewpoints.


Our proposed approach uses a novel S-BEV map which can be easily augmented to introduce diverse viewpoints in angle, longitudinal, or lateral change. It also incorporates a coarse to fine approach unlike some approaches that use an intermediary image retrieval \cite{irschara2009pcls, sarlin2019coarse} which allows us to learn features of specific parts of the map in an auxiliary way.

\subsection{Motion Estimation while Learning Depth}
There has been a lot of reent work of estimating depth \cite{eigen2014depth} from a single image. Some of the most recent successes with this have been with monodepth2 \cite{monodepth2} and packnet-sfm \cite{Guizilini2019PackNetSfM3P} which use separate techniques in treating depth estimation as an auxiliary task and using 3D packing to generate accurate depth maps. Specifically, monodepth2 trains a pose estimation network between frames to eventually generate accurate depth maps.

DeMon \cite{DeMoN} and UndeepVO \cite{undeepVo} are also two deep learning based approaches that are able to do the pair task of motion estimation through an $[R | t]$ matrix and absolute scale depth estimation from a stream of images.

Our approach uses stereo for depth estimation and to unproject the semantic 2D image into 3D space for bird's eye view representation, but could use mono-depth techniques for this.

\subsection{Robust Coarse-to-Fine Localization}
Due to the compute-intensiveness of directly predicting pose from an SfM model, many approaches have tried to do coarse-to-fine localization where they first narrow down to candidate locations from a global search and only then match local image based features. While most approaches in the past have used local features such as SIFT \cite{SIFT}, newer state-of-the-art approaches like HF-Net \cite{sarlin2019coarse} use more robust local features such as deep-learning based SuperPoint \cite{superpoint} which seem to be consistently outperforming other traditional local features. Specifically, HF-Net is able to leverage these learned descriptors to perform robust localization across a large set of changes in appearance.

Another coarse-to-fine approach is one using a topo-metric map \cite{roussel2020localization} where one first uses a traditional SLAM approach to create a map along with paired sensor data including monocular or stereo images. This map is divided into discrete topological nodes which are defined by a specific pose, created based on constant translational and angular distances between each other. Models first localize to a specific node in the map and then find their positions relative to those nodes using traditional techniques like Perspective-n-Point (PnP).

While both coarse-to-fine approaches provide reasonable pose estimates, they lack the ability to perform well in drastically varying weather and lighting conditions such as rain, snow, sunset, fog, etc.

Framing our problem in a coarse-to-fine setting, however, is advantageous to us as we use the S-BEV representation which has the potential to account for varying lighting and weather conditions, and is then used to create embeddings that are used for both the course and fine localization.

S-BEV localization which accounts for these varying weather and lighting conditions as we will later address.

\subsection{Semantic Localization}
Semantic localization describes any approach that uses semantic labels (car, vegetation, building, sign, etc) to aid itself in localizing within a scene. These approaches tend to mostly be weather invariant considering that semantic labels don't change across seasons.

Schönberger et. al \cite{schonberger2018semantic} proposes an approach based on a full 3D semantic reconstruction of the world where a generative model is used to create a full scene from a partial understanding of a scene. Since the representation is 3D, it accounts for viewpoint changes in which prior 2D approaches fail. Traditional descriptors such as SIFT \cite{SIFT} or deep-learning based descriptors such as SuperPoint \cite{superpoint} aren't used given that they aren't traditional RGB images. The latent representation created by this generative model is then used for localization.

Stenborg et. al \cite{stenborg2018semantic} proposes a slightly different which is able to utilize SIFT-based features. This approach uses the fact that SIFT works on images which span a short range of time given that the lighting or weather isn't drastically different. Using these features, points can be consistently projected from a 3D map into a 2D semantic image and then be evaluated based on how many points of a certain semantic class lie within a specific semantic label. A particle filter is then weighted based on these evaluations to localize.

Garg et. al \cite{Garg_2020} used \textit{Delta Descriptors}, which is a change-based representation and is robust to appearance variation during a revisit. Furthermore, they are able to achieve state-of-the-art performance by combining these representations with sequence-based matching. Other image signatures such as the horizon line \cite{ho2014localization} has been used to relocalize vehicle on freeways by first detecting a 1-D horizon line signature and then using a particle filter which in-turn uses lane marking patterns and their frequency to build a motion model for robust localization.

While these approaches both account for changes in weather and lighting, they do not account for significant structural changes that can occur across seasons such as mounds of snow or change in vegetation. Using a birds-eye view map is favorable in this sense.

\section{METHOD}
\label{methodMarker}

Our method of localization relies on encoding a database of various locations from a map into a deep learning model. This is achieved by first training an autoencoder to predict generic scene characteristics from a single S-BEV view of the scene. Once this has been done, we generate latent embeddings using just the encoder and feed this in both to a nearest neighbor classifier a separate fully connected network for relative pose prediction. A high-level architecture is detailed in Figure \ref{fig:highLevelArch0}.

\subsection{Topo-metric Map Construction}
\label{eqMarker}

We first associate global camera poses with each of our stereo camera pairs. For indoor environments this can be done with an off-the-shelf method such as ORB-SLAM2 to extract trajectory. For larger-scale environments, we can use IMU-GNSS EKF-based approaches to extract camera poses. Following this, we must mark topological nodes via global camera poses. Nodes are marked on map-by-map basis where we select translational and angular thresholds as described in \cite{roussel2020localization}.

\subsection{Semantic Bird’s Eye View Map Construction}
To create a Semantic Bird’s Eye View Map (S-BEV), we require semantic segmentation maps as well as disparity maps for our stereo pair. We use StereoSGBM with a Weighted Least Squares (WLS) Filter for post-filtering to create a smooth disparity map from a pair of rectified stereo images which we convert to real-word distances via camera intrinsics.

We run an off-the-shelf semantic segmentation network on the left image of our stereo pair. We use ResNet50dilated + PPM-deepsup as implemented with 150 classes as trained with ADE20K dataset by MIT CSAIL \cite{zhou2017scene, zhou2018semantic}, consisting of both indoor and outdoor classes. However, we narrow to 30 classes that only show indoor. While the segmentation network can be finetuned to fewer classes, we show that any off-the-shelf network provides good results. We also choose to ignore any classes associated with dynamic objects like "cars" or regions that cause inconsistent depth like "sky".

Using the segmentation output along with the depth map, we project our image into a 3D point cloud based on camera intrinsics. We then rotate this point cloud to view from above. (Figure \ref{fig:overview}). We project multiple stereo pairs in the same map, compensating for motion via camera extrinsics to encourage greater stability and less change across S-BEV map frames. Specifically, we plot the current plus the last four stereo pairs into one motion-compensated S-BEV map. Our S-BEV is a 352x352 sized image which is then used throughout the rest of our pipeline.

\begin{figure}[t]
\begin{center}
    \includegraphics[width=0.75\linewidth]{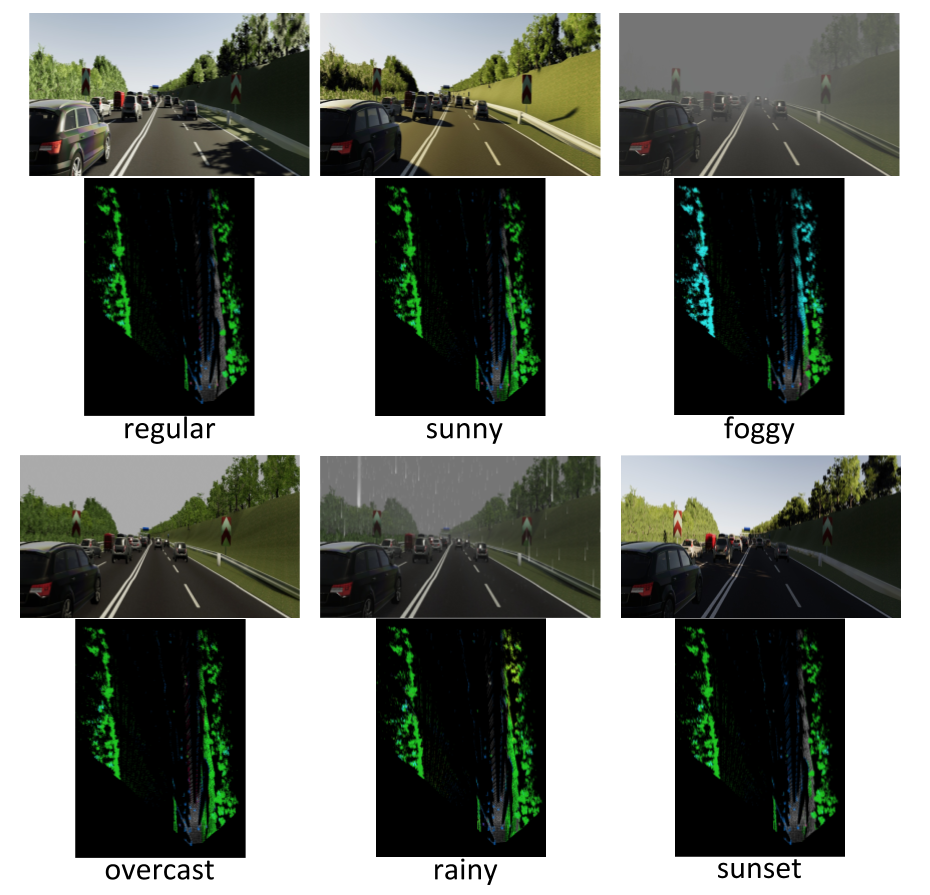}
\end{center}
\vspace*{-0.35cm} 
   \caption{Sample S-BEV constructions on six different weather conditions on Scene 20 of the vKITTI dataset. Despite the change in weather and lighting conditions, the structure in the S-BEV signature remains similar, illustrating its potential for localization across weather and lighting conditions.}
\label{fig:sbevMultiWeather}
\vspace*{-0.45cm} 
\end{figure}

As seen in Figure \ref{fig:sbevMultiWeather}, our S-BEV map stays relatively constant across the six varying weather conditions (in the vKITTI dataset) shown. The main discrepancy is in foggy weather where trees toward the back of the scene are misclassified. However, because the class IDs outputted by our segmentation network are arranged in such a way that similar classes are numerically close to each other, this misclassification won’t affect the rest of our pipeline. Also, the structure of the scene stays constant throughout all the images which sheds light on the factors which went into choosing S-BEV for our pipeline. However, this issue can also be mitigated through fine-tuned segmentation.

This highlights the fact that regardless of the weather and lighting condition, our input image will be processed in the exact same way through the rest of our pipeline due to the conversion into S-BEV.

\subsection{Topo-metric Node Localization}
Using the S-BEV representation and an autoencoder paired with a nearest neighbors classifier, we perform a coarse localization to topological nodes.

\begin{figure*}[t]
\begin{center}
    \includegraphics[width=0.98\linewidth]{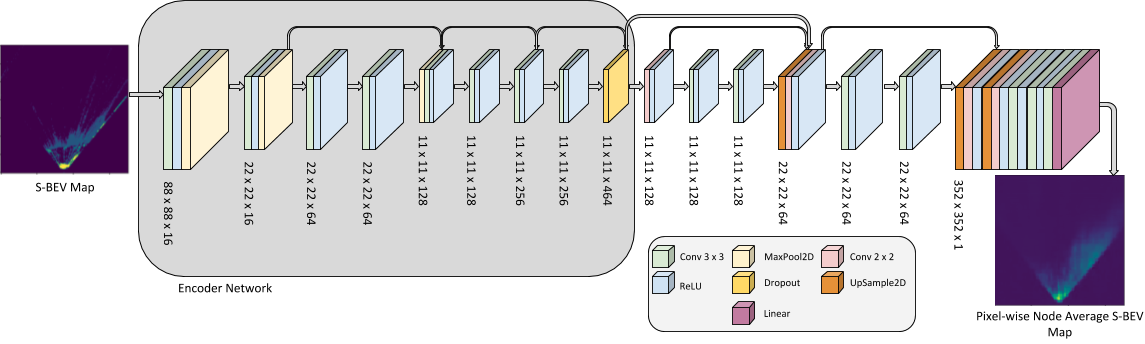}
\end{center}
\vspace*{-0.35cm} 
   \caption{Autoencoder architecture. Encoder portion in gray box is used to generate latent embeddings of S-BEV map.}
\label{fig:aeArch}
\vspace*{-0.45cm} 
\end{figure*}

We first train an autoencoder using the S-BEVs we’ve generated for all the training images. The architecture (see Figure \ref{fig:aeArch}) used consists of convolution filters of varying kernel sizes paired with skip connections between a few layers. However, rather than reconstructing the input BEV image, our autoencoder reconstructs a BEV image representing the pixel-wise average of all S-BEVs for a specific node. The pixel-wise average is meant to represent generic characteristics of an entire node scene while ignoring things such as dynamic obstacles which don’t play a role in understanding our location within a scene. This encourages our network to learn its translational and rotational fit in relation to the entire node region rather than trying to reconstruct holes possibly generated by long-term artifacts such as snow mounds or dynamic objects such as cars and trucks. The network is trained with MSE loss.

We then detach the decoder part of our autoencoder and pass a candidate S-BEV map through the encoder to generate a latent embedding. We then take our latent embedding and perform a nearest neighbors classification to other sample latent embeddings from topological nodes across the entire map. The node of the numerically closest embedding based on Euclidean distance is our node classification. Our ground truth data is generated by finding the node with the minimal topo-metric distance to our sample.

\subsection{Fine Localization using 3-DoF Regressor}
\label{relPoseEq}
We perform fine localization of relative 3-DoF pose (x, y, theta) to the topological node using a 3-layer fully connected network with our input being a concatenated version of our one-hot encoded node classification and flattened latent embedding. We also add multiple layers of dropout to prevent overfitting.

Ground truth data being generated for this regressor model is based on pose relative to our topological node. Known variables include closest node global pose and camera global pose. The X, Y, and theta are calculated as follows: $Pose_{rel} = Pose_{cn}^{-1} \times Pose_{cf}$.

where $\mathcal Pose_{rel}$ is the pose of the current frame relative to the closest topological node, $\mathcal Pose_{cn}$ is the global pose of the closest node, and $\mathcal Pose_{cf}$ is the global pose of the current frame. This is often confused with multiple coordinate frames which exist within our scene including global X and Y as well as X and Y relative to our ground truth pose at that specific timestamp. However, neither of these are the same as our pose relative to the closest topological node which is an arbitrary distance in front of or behind the candidate pose.

\subsection{Final Localization Output}
After we obtain relative 3-DoF pose as well as the topological node classification we can combine them by doing the inverse operation of what was described in "\nameref{relPoseEq}" as follows with the same variable names and labels: $Pose_{cf} = Pose_{cn} \times Pose_{rel}$.

\subsection{Localization Filter}

Once we have the final localization output from the 3-DoF regressor in a global frame, we use a Kalman Filter (KF) framework for fusing that with the GPS/IMU output. The state vector is represented by \begin{math} \mu = [x, y, \theta] \end{math}. The prediction step involves predicting the motion of the vehicle between two iterations of the filter. We obtain velocities from the Applanix POS LV\cite{applanix} and apply a constant velocity motion model to estimate the pose of the vehicle (also called predicted state).
The update step is where the state of the filter is updated or corrected using measurements obtained from the fine 3-DoF regressor. These measurements correct the vehicle pose in \(x\), \(y\) and \(\theta\). Mathematically, this fusion is performed via the iterative localization updates

 \begin{alignat}{2} \label{EKF}
	\text{Predict:} & \quad \bar{\mu}_k = F_{k-1} \mu_{k-1} \\
	& \quad \bar{\Sigma}_k = F_{k-1} \Sigma_{k-1} F_{k-1}^T + Q_{k-1} \nonumber \\
	\nonumber \text{Update:} & \quad K_k = \bar{\Sigma}_k H^T_k ( H_k \bar{\Sigma}_k H^T_k + R_k)^{-1} \\
	\nonumber & \quad \mu_k =  \bar{\mu}_k + K_k(z_k - h_k( \bar{\mu}_k )) \\
	\nonumber & \Sigma_k = (I - K_k H_k) \bar{\Sigma}_k (I - K_k H_k)^T + K_k R_k K_k^T
 \end{alignat}
 
where $F_k$ represents the motion model of the vehicle and $Q_k$ is the corresponding uncertainty, $z_k$ is the output of the fine 3-DoF regressor and $R_k$ is the corresponding uncertainty estimated as a fit to the covariance of the 3-DoF regressor algorithm. We use GPS to initialize the filter in the linearized global frame which results in high uncertainty for the first few measurements.

\section{EXPERIMENTS}
We use two datasets for our experiments, the vKITTI2 \cite{gaidon2016virtual, cabon2020vkitti2} and the Ford AV \cite{agarwal2020ford} datasets. vKITTI2 is a synthetic dataset that contains virtual clones of 5 sequences from the original KITTI dataset. We use scene 20, which at 1.2 km in length, is the longest sequence to test the performance of our S-BEV signature across different weather and lighting conditions. The Ford AV dataset is a real-world dataset collected from 3 vehicles that drive average route lengths of 66 km through freeway and suburban Michigan over a period of 2 years. We use log 1 from vehicles V2 and V3, collected on 2017-08-04. This is primarily a freeway scene, with occasional overpasses, under cloudy conditions. We use the provided front-left and front-right camera images for stereo depth generation, along with global pose information provided to create the topological nodes. Topological node thresholds are 20m and $30 deg$ for vKITTI2 and 40m and $30 deg$ for the Ford AV dataset.


We first evaluate the car's ability to localize itself within the same run of the same path taken during data collection (rows 1-3 of Table \ref{tab:stat_weather}). Images are first categorized as belonging to a topological node (based on real-world distance to closest topo-node). Subsequently, for each topo-node, images are split into train and test using a random 80-20 split. Thus training images get slightly different viewpoints relative to the test images even though they are collected from the same run. The training images are then used the generate the depth maps, the segmentation images, and finally the S-BEV images. S-BEV images are used to train the auto-encoder, the nearest neighbour classifier and the 3-DoF models. 

\begin{figure}[t]
\begin{center}
    \includegraphics[width=0.98\linewidth]{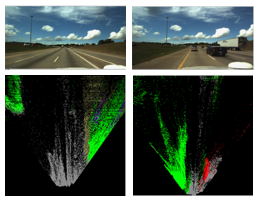}
\end{center}
\vspace*{-0.35cm} 
   \caption{S-BEV from the same topological node from V2 and V3. These vehicles are travelling in different lanes along the same route. This leads to considerably different S-BEVs along different lanes at the same location on the freeway, leading to a decrease in accuracy of our current method.}
\label{fig:SBEVdiff}
\vspace*{-0.5cm} 
\end{figure}

Subsequently, we consider the car's (V3's) ability to localize itself within a map made by another vehicle (V2) while travelling along the same route (rows 4 of Table \ref{tab:stat_weather}). Training images are now from a completely different perspective from the test images as the cars travel along different lanes (\ref{fig:SBEVdiff}.

We evaluate localization accuracy with 3 metrics. The first is node accuracy, which is determined by whether the coarse localizer (nearest neighbour classifier) localizes to the correct closest node. We then present $(x, y, theta)$ accuracies which describe the MAE error in our fine 3-DoF localizer, with respect to its closest node. Finally, we append an EKF to the pipeline, that combines the S-BEV localization output with a GPS/IMU, after which we have filtered $(x, y, theta)$ accuracies. The coordinate system of our ego vehicle is aligned in such a way where X is in the direction of travel, Y is to the left of this, and Z is pointing towards the sky.


We have performed an ablation study (Table \ref{tab:ablation}) by evaluating the efficacy of some of the techniques we used in training the Auto-encoder. $BASE$ represents the full approach described in the Method section. $AVG$ describes an ablation when the AE is trained to reconstruct the exact S-BEV input and not the smoothed/averaged version of the S-BEV over all the images in a topo-node. Finally, $AUG$ removes the angular or off-by-one data augmentations described earlier.



\begin{figure}[h]
	\centering
	\captionsetup{justification=centering}
	\subfloat{\includegraphics[width=0.95\linewidth, trim=0 0 0 0, clip]{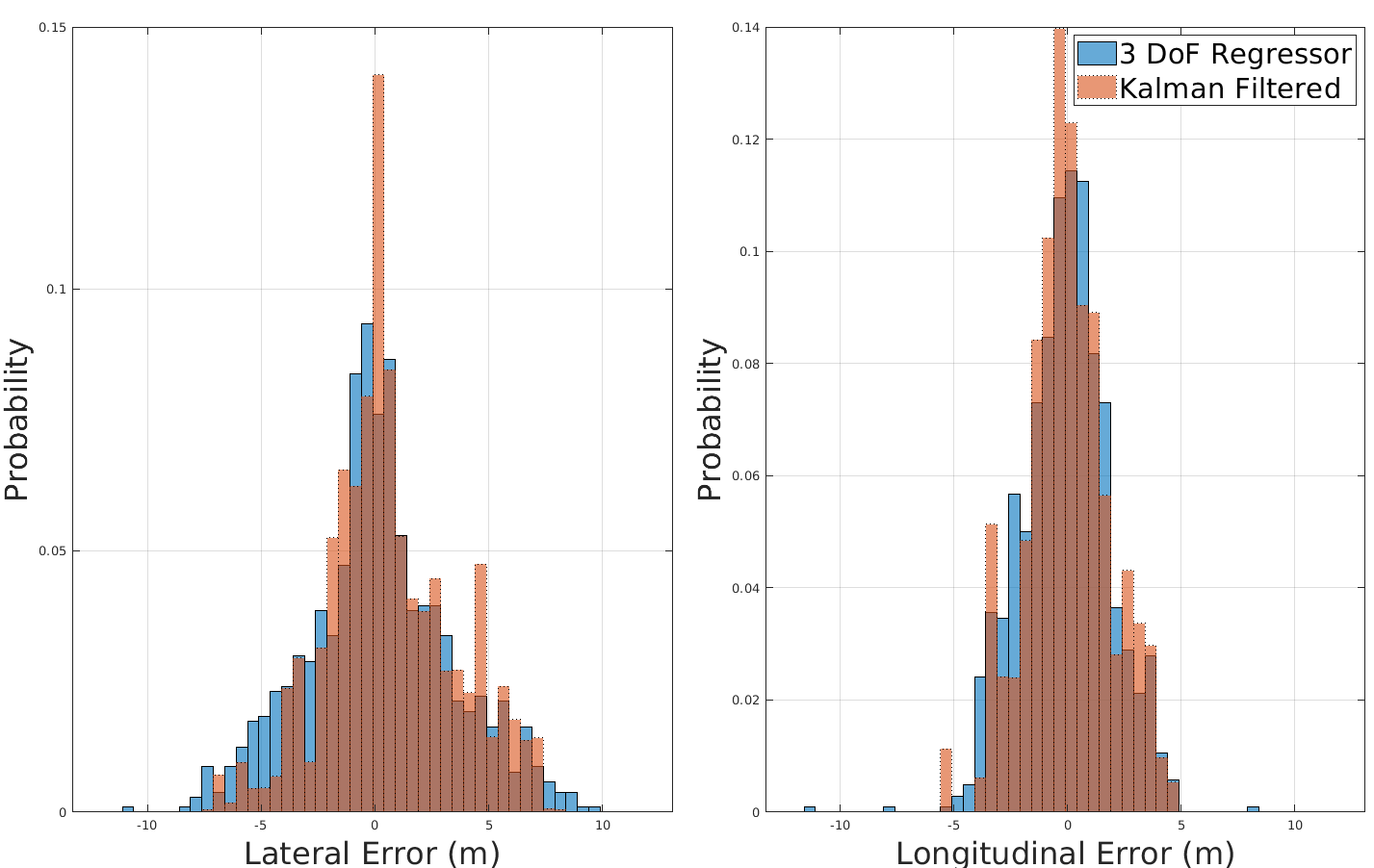}}
	\caption{Localization error comparisons for V2}
\label{fig:LocError}
\vspace*{-0.5cm} 
\end{figure}

Table \ref{tab:stat_weather} demonstrates the results of using the full $BASE$ model on vehicles V2 \& V3 (Log 1) of the Ford AV dataset. Along the columns, we have the performance of the nearest neighbour based course localizer ($acc_{\textbf{node}}$) and then the performance of the fine-grained 3-DoF localizer to the node.
The nearest neighbour localization accuracy results show the percentage of test data that report the correct closest topological node. The 3-DoF localization errors are the x, y and theta errors relative to the node. In this table, we assume perfect nodal localization accuracy for the 3-DoF localizer, but the following EKF results (Figure \ref{fig:LocError}) shows the accuracy of the full pipeline, including nodal accuracy. Along the rows, we have results from the vKITTI and Ford AV datasets. V2 and V3 represent in-route localization errors for the same vehicle, tested on left out parts of the dataset.  V3$\xrightarrow[]{}$V2 shows results for a model trained on V3 and tested on V2. Figure \ref{fig:SBEVdiff} shows the difficulties in matching a V2 to a V3 trajectory. The vehicles travel in different lanes, resulting in different views of the grass banks on either side of the highway. At the same point in the route, V2 travels in the right-most lane, which results in a big grass patch to the right, in the S-BEV signature; while V3 travels in the left-most lane, resulting in the grass patch shifting to the left. This results in our simple nearest neighbour localizer not working at all for this cross-vehicle case, but the 3-DoF localizer (assuming perfect topo-node accuracy) still reports reasonable results.

Next, we append the pipeline with the EKF described earlier, and report before and after filtering accuracy. These pre-EKF numbers include the error from the topo-node localization (and are hence different from Table \ref{tab:stat_weather}). Figure \ref{fig:LocError} shows the error distribution in the longitudinal (X) and lateral (Y) directions for V2. The 3-DoF regressor accuracy (\begin{math} MAE_{lat} = 2.68m, MAE_{long} = 1.7m \end{math}) improves when fused with GPS/IMU using a Kalman Filter (\begin{math} MAE_{lat} = 2.17m, MAE_{long} = 1.43m \end{math}). We see a similar behaviour in case of V3 where the 3-DoF regressor accuracy (\begin{math} MAE_{lat} = 3.4m, MAE_{long} = 3.55m \end{math}) is improved by filtering it (\begin{math} MAE_{lat} = 2.4m, MAE_{long} = 2.55m \end{math}).

We also do a cross-weather localization study using the virtual vKITTI dataset. Table \ref{tab:weather_variation} illustrates the effect of different weather conditions along the same route on the course and fine localization accuracies. In the virtual dataset atleast, changing visibility and lighting does not affect the localization accuracy, which is constant across weather change. 

The current version of the sytem uses a relatively simple nearest neighbour classifier for classifying an S-BEV embedding to a topo-node. This is not effective for perspective changes caused while travelling in different lanes (Figure \ref{fig:SBEVdiff}). We hope to improve the course classifier by training it with a contrastive loss as in a Siamese network that will force the embeddings of the S-BEVs from the same topo-node to be close to each other and the embeddings from different topo-nodes to be far from each other. We could also train a network to generate a cleaner S-BEV from the forward perspective camera images \cite{ng2020bev}. We expect these approaches to be more robust to lane changes and real-world lighting \& weather changes and leave this as future work.

\begin{figure}[t]
\begin{center}
    \includegraphics[width=0.98\linewidth]{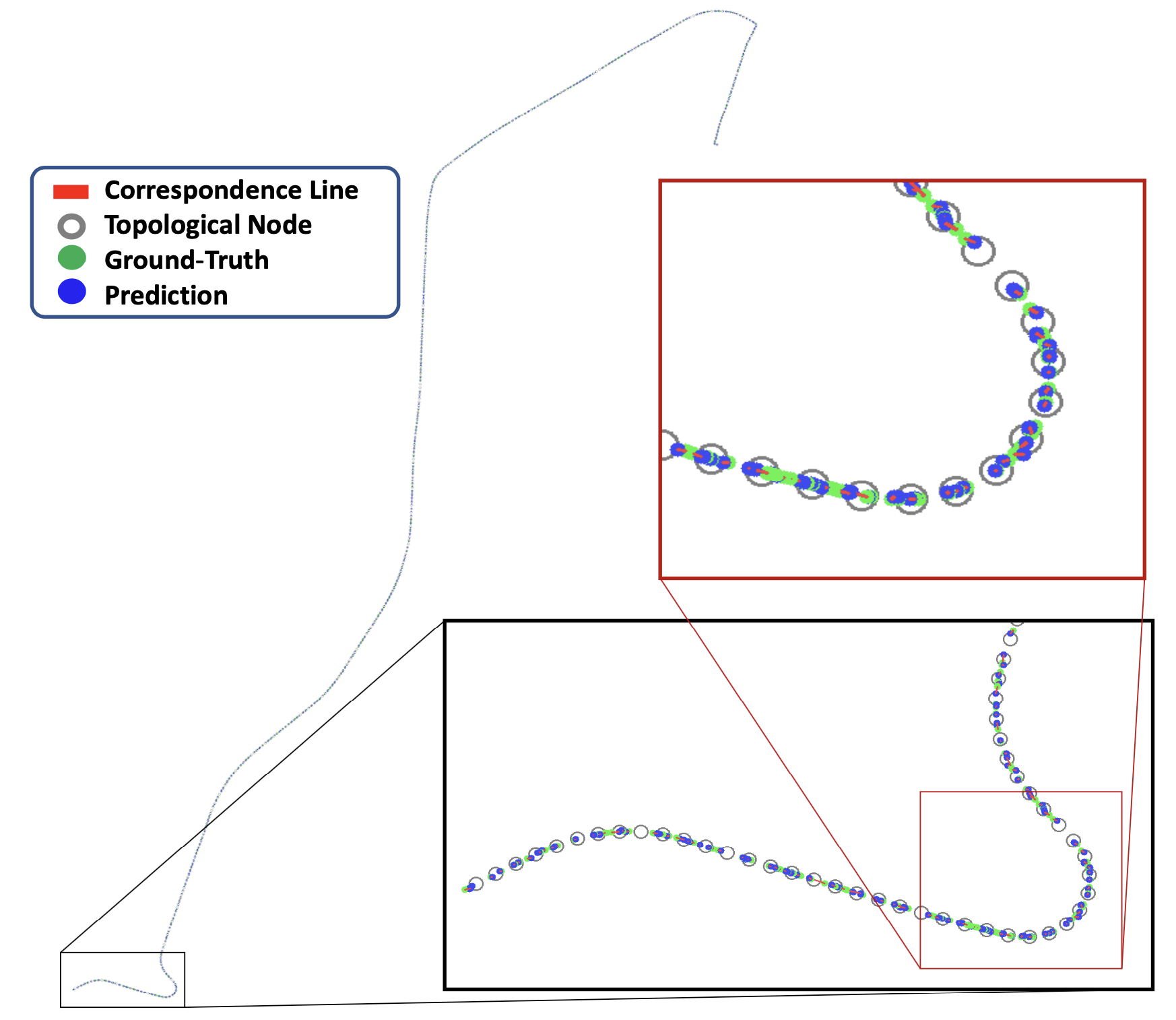}
\end{center}
\vspace*{-0.35cm} 
   \caption{Course and fine localization results for the 22 km trajectory in Log 1, V2 of the Ford AV dataset.}
\label{fig:globalPath}
\vspace*{-0.7cm} 
\end{figure}

\begin{table}[!htb]
    \begin{center}
      \caption{\textbf{Ablation Tests for Node Accuracy}}
      \begin{tabular}{llll}
      \toprule
        \textbf{Dataset} & BASE & AVG & $AUG$ \\
        \hline
        vKITTI 2    &   99.5    &   96.6   &   96.5 \\
        Ford AV V2      &   98.7    &   90.1   &   88.4 \\ 
        \hline
        \label{tab:ablation}
      \end{tabular}
  \end{center}
  \vspace*{-1.05cm} 
\end{table}

      

\begin{table}[!htb]
    \begin{center}
      \caption{\textbf{Static Weather Results}}
      
      \begin{tabular}{lllll}
      \toprule
        \textbf{Dataset}&$acc_{\textbf{node}}$&$err_{\textbf{x}}$ (m) &$err_{{\textbf{y}}}$ (m)  &$err_{\textbf{theta}}$ (deg) \\
        \hline
        vKITTI 2    &   99.5    &   0.27    &   0.04   &   0.2 \\
        Ford AV V2      &   98.7    &   1.50    &   0.795   &   1.2 \\
        Ford AV V3      &   94.7    &   3.79   &   1.953   &    1.4 \\
        Ford AV V3$\xrightarrow[]{}$V2 &   -    &   4.82   &   2.46   &   1.59 \\
        \hline
        \label{tab:stat_weather}
      \end{tabular}
  \end{center}
    \vspace*{-1.05cm} 
\end{table}

\begin{table}[!htb]
    \begin{center}
      \caption{\textbf{Weather Variation Results}}
      
      \begin{tabular}{llllll}
      \toprule
        \textbf{Dataset}&\textbf{Weather}&$acc_{\textbf{node}}$& \shortstack{$err_{\textbf{x}}$ \\ (m)} & \shortstack{$err_{{\textbf{y}}}$ \\ (m)} & \shortstack{$err_{\textbf{theta}}$ \\ (deg)} \\
        \hline
        vKITTI 2    &   Regular   &   99.5     &   0.27   &   0.04  &   0.2 \\
        vKITTI 2    &   Sunny     &   99.4     &   0.22   &   0.06  &   0.2 \\
        vKITTI 2    &   Foggy     &   99.1     &   0.23   &   0.05  &   0.4 \\
        vKITTI 2    &   Overcast  &   99.3     &   0.30   &   0.07  &   0.5 \\
        vKITTI 2    &   Rainy     &   99.5     &   0.22   &   0.04  &   0.3 \\
        vKITTI 2    &   Sunset    &   99.0     &   0.27   &   0.05  &   0.2 \\
        \hline
        \label{tab:weather_variation}
      \end{tabular}
  \end{center}
    \vspace*{-1.05cm} 
\end{table}

\section{CONCLUSIONS}
We describe a Semantic Bird's Eye View (S-BEV) representation generated from a forward facing stereo camera that allows a vehicle to be localized along a pre-mapped route. We conduct experiments on the vKITTI2 virtual dataset with MAE of 0.27m \& 0.04m in the longitudinal and lateral directions and show that the S-BEV representation is robust to weather and lighting changes. We also demonstrate MAE of \textbf{2.17m} on a 22 km long highway route in the Ford AV dataset. Through these results, we can see that the SBEV representation has good potential for vision based localization in dynamically changing environments.



\FloatBarrier

\addtolength{\textheight}{-4cm}   





\bibliographystyle{IEEEtran}


\bibliography{sbev}

\end{document}